\documentclass[10pt,journal,compsoc]{IEEEtran}
\ifCLASSOPTIONcompsoc
  \usepackage[nocompress]{cite}
\else
  \usepackage{cite}
\fi

\ifCLASSINFOpdf
\else
\fi

\usepackage{amsmath}

\usepackage{multirow}
\usepackage{multicol}
\usepackage{amsmath}
\usepackage{mathtools}
\usepackage{graphicx}
\usepackage[dvipsnames]{xcolor}
\usepackage{tablefootnote}
\usepackage{comment}

\usepackage{url} 
\usepackage[colorlinks=true, allcolors=blue]{hyperref}

\begin{document}
%
\title{Global Attention-based Encoder-Decoder LSTM Model for Temperature Prediction of Permanent Magnet Synchronous Motors}

\author{Jun~Li,
        Thangarajah~Akilan,~\IEEEmembership{Member,~IEEE}
\IEEEcompsocitemizethanks{\IEEEcompsocthanksitem J. Li is with the Department of Electrical and Computer Engineering, Lakehead University, Thunder Bay, Canada.\protect\\
E-mail: jli12363@lakeheadu.ca
\IEEEcompsocthanksitem T. Akilan is with the Department of Software Engineering, Lakehead University, Thunder Bay, Canada.\protect\\
E-mail: takilan@lakeheadu.ca}

}

%
%

\markboth{}%
{Shell \MakeLowercase{\textit{et al.}}: Bare Demo of IEEEtran.cls for Computer Society Journals}



\IEEEtitleabstractindextext{%
\begin{abstract}
Temperature monitoring is critical for electrical motors to determine if device protection measures should be executed. However, the complexity of the internal structure of Permanent Magnet Synchronous Motors (PMSM) makes the direct temperature measurement of the internal components difficult. This work pragmatically develops three deep learning models to estimate the PMSMs' internal temperature based on readily measurable external quantities. The proposed supervised learning models exploit Long Short-Term Memory (LSTM) modules, bidirectional LSTM, and attention mechanism to form encoder-decoder structures to predict simultaneously the temperatures of the stator winding, tooth, yoke, and permanent magnet. Experiments were conducted in an exhaustive manner on a benchmark dataset to verify the proposed models' performances. The comparative analysis shows that the proposed global attention-based encoder-decoder (EnDec) model provides a competitive overall performance of 1.72 Mean Squared Error (MSE) and 5.34 Mean Absolute Error (MAE).
\end{abstract}

}

\maketitle

\IEEEdisplaynontitleabstractindextext

%
\IEEEpeerreviewmaketitle

\IEEEraisesectionheading{\section{Introduction}\label{sec:introduction}}

\begin{figure}[!t]
    \centering
    \includegraphics[width=0.8\columnwidth]{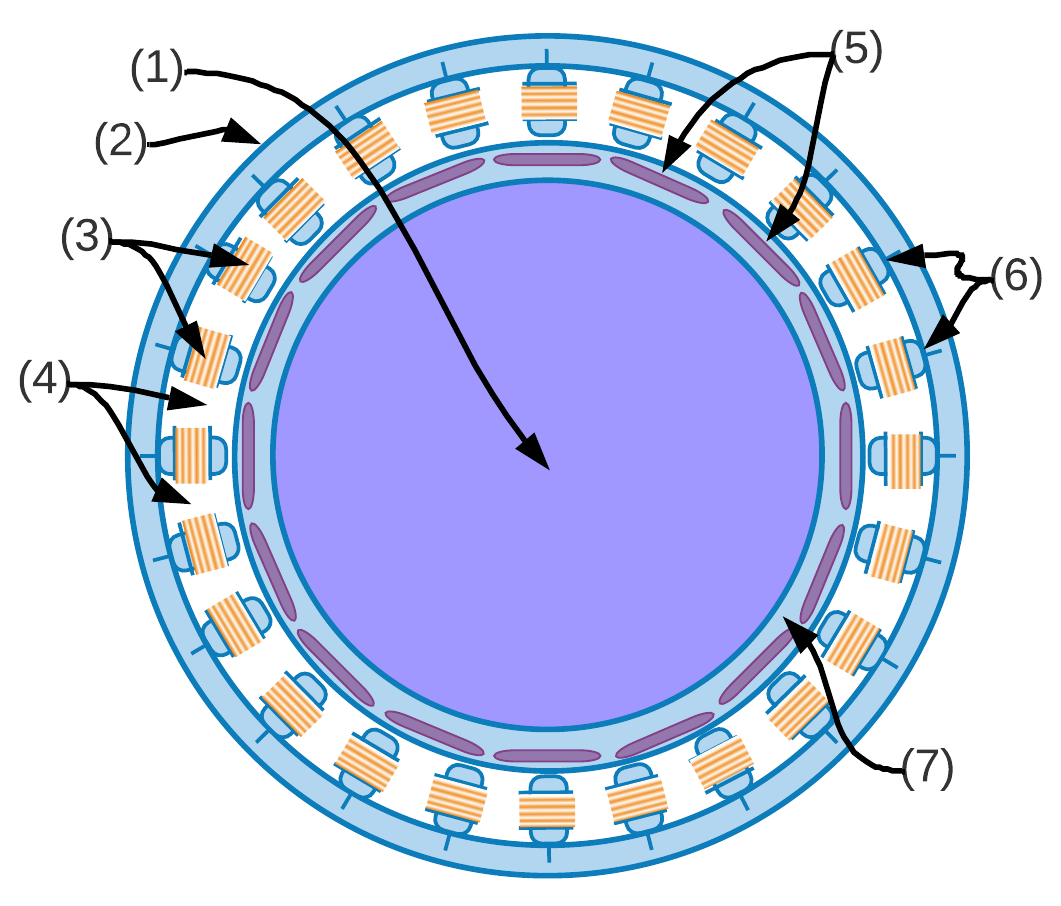}
    \caption{Illustration of the cross section of PMSM: (1) rotor shaft, (2) stator yoke, (3) stator winding, (4) stator slot, (5) permanent magnet, (6) stator tooth, and (7) rotor ion.}
    \label{fig:pmsm}
\end{figure}

\begin{figure*}[!htp]
\centering
\includegraphics[trim={2.15cm, 2.7cm, 3.1cm, 1.5cm}, clip, width=0.99\textwidth]{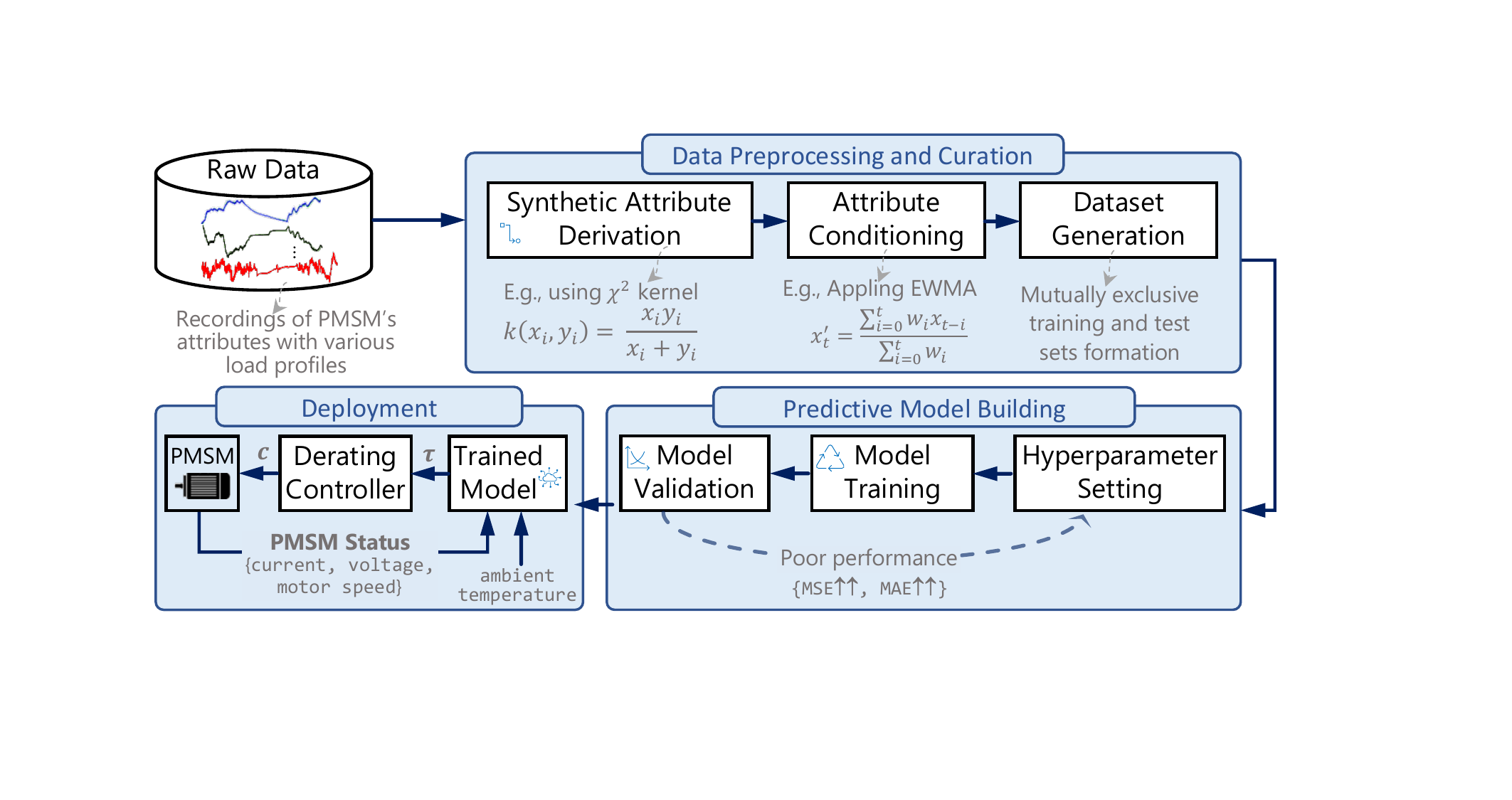}
\caption{Overview of the proposed PMSM's thermal stress prediction and illustration of a model deployment scenario: \textbf{$\tau$} - vector of predicted temperatures of stator winding, stator tooth, stator yoke, and permanent magnet; \textbf{$c$} - derating control signal that regulates PMSM operation.}
\vspace{-0.2cm}
\label{fig_process_flow}
\end{figure*}

\IEEEPARstart{W}{ith} ever-increasing awareness for green energy, the development of electric vehicles (EVs) has become a mainstream product in the automotive industry. PMSMs in EVs have been widely adopted for a range of automotive motor applications, due to their excellent characteristics, viz. large power density and torque, low noise and vibration, high reliability and efficiency, and quicker dynamic response. 
However, to fully take advantage of these characteristics, it is vital to early detect the thermal stress on the PMSMs to prevent any component failure. For instance, excessive thermal stress can cause the insulation varnish of the stator winding to melt and the permanent magnets to become irreversibly demagnetized~\cite{kirchgassner2019deep}, resulting in reduced life-time of the PMSMs~\cite{kirchgassner2019deep, wallscheid2015global}. 
Although a sensor-based technology can be used for the temperature measurement of the PMSMs, it is not applicable for some conditions. For example, placement of a thermal sensor on the rotor has low engineering feasibility due to the complex and confined internal structure of the PMSMs as illustrated in Fig.~\ref{fig:pmsm}. Besides, these sensors have a lower lifespan than the components of the PMSMs, i.e., the functionality of the thermal sensors deteriorates faster than the PMSM's components. Thus, there is a demand for research and development of data-driven approaches that can precisely estimate the different components' temperature of the PMSMs using externally measurable variables, like the voltage and current in real-time. These temperature estimations can be used as the input parameters of intelligent derating controllers (cf.~Fig.~\ref{fig_process_flow}) that guarantee safe operations, while maximizing the runtime utilization of the motors. 

The temperature prediction of PMSMs is a sequence data analytical problem; more specifically, it is a time-series regression analysis. To tackle with this, over the past decade, there were several approaches proposed from pure mathematical modelling to advanced data-driven supervised learning systems. For instance, the earlier approaches use electrical machine modelling with precise flux observer or signal injection techniques, computational fluid dynamics (CFD), finite element analysis (FEA), and heat transfer computation via equivalent circuit theories~\cite{wallscheid2015global, reigosa2010magnet, zhou2020rotor}. These approaches demand comprehensive domain knowledge, and precise selection of parameters. Due to these challenges, they are not widely adopted by the industrial community. 

On the other hand, the data-driven approaches collect representative data samples from an apparatus, then train machine learning or deep learning models on the collected samples. For example, Kirchgässner and Wallscheid~\cite{kirchgassner2020estimating} setup a data acquisition test bench with a 3-phase PMSM (52 kW) and embedded thermocouples to collect measurements. They recorded all key quantities representing various load profiles through a dSPACE DS1006MC rapid-control-prototyping system\footnote[1]{\url{https://www.dspace.com/en/pub/home/news/dspace-rapid-prototyping-syste.cfm?nv=nb}} (for further details about this data acquisition, one can refer to \cite{kirchgassner2020estimating}). 
In this line, many solutions have been proposed from basic linear regression models to sophisticated deep neural networks (DNNs). From the literature, it is found that the basic regression models are not robust for PMSMs' high-fluctuating thermo-temporal properties and do not produce accurate results. With the advancement of Deep Learning (DL) technology, the models developed using recurrent architectures, viz. LSTM units, Gated Recurrent Units (GRUs), and Temporal Convolutional Network (TCN) demonstrate robustness and high-level precision for various time-series analytical problems~\cite{akilan2019video, kirchgassner2019deep, kirchgassner2020estimating, wallscheid2017investigation, 8671459, chung2014empirical, farha2019ms}. 

Thus, this work aims to systematically investigate, develop, and verify more robust DL models by exploiting bidirectional sequence-to-sequence LSTMs, and global attention mechanisms for accurate predictions of the temperatures of key components inside PMSMs, including stator components and permanent magnet using input quantities, such as, ambient and coolant temperatures, supply current and voltage, and motor speed. Hence, in this work we build the models with increasing complexity from a baseline encoder-decoder LSTM (EnDec-LSTM) to bidirectional EnDec-LSTM, and global attention-based EnDec-LSTM. We also conduct exhaustive ablation studies to analyze the effectiveness of the proposed architectures. 

The rest of this article is organized as follows. Section~\ref{literature-review} reviews related works, Section~\ref{methodoloy} elaborates the proposed recurrent models, Section~\ref{experimental-results} discuss the experimental findings, while Section~\ref{conclusion} concludes the work with future directions.


\section{Related Work}\label{literature-review}

\subsection{Sequence Data Analytical Models}\label{seq-modeling}

The conventional DL solutions for sequence data analysis are based on Recurrent Neural Networks (RNNs)~\cite{montgomery2015introduction, 8818627, cho2014learning}. For instance, Cho~\textit{et al.}~\cite{cho2014learning}, introduced an Encoder-Decoder (EnDec) RNN architecture for translating phrases from one language to another language.  
However, the vanilla RNNs face a major problem of vanish gradient due to repeated recurrent connections and recursive derivative computation for backpropagation through time (BTT) during training. This problem makes the RNNs difficult to be trained on data with long-term dependencies, like the data in this work. 

Some researchers introduced a bidirectional training strategy, to enhance the standard RNN's ability to handle long-term dependent data sequences. 
For example, Schuster and  Paliwal~\cite{schuster1997bidirectional} generalized the causal structure of the RNNs to a non-causal learning model, coined bidirectional recurrent
neural network (BRNN). The causal RNNs' output at time stamp, $t$ does not depend on future values of inputs, but the BRNN's response at time stamp, $t$ depends on both the past and the future input values~\cite{ 9340579, 1333842}. It is notable that when the BRNN hidden layers structured with LSTM units, the resulting architecture will be a Bidirectional LSTM (BLSTM). Hence, forming DNNs by stacking of several hidden layers of BRNN or BLTM can produce better generalized performance~\cite{8818627, del2018speaker, 8845643}.

To further improve the sequence learning ability of the recurrent networks, attention mechanisms have been integrated with EnDec RNNs as in~\cite{bahdanau2014neural, luong2015effective, cheng2016long}. The conventional EnDec RNNs work on a fixed length context vector, thus there is a chance for forgetting key information conveyed in long input sequences. However, the improved attention-based models generate the context vector representing the relationship between input and output sequences using three elements: encoder hidden states, decoder hidden states, and alignment information between input and target (cf.~\ref{fig:attention-lstm_cell}). Through this approach, these models 
make sure all past contextual information are not forgotten during the target prediction in every time step. Similar to the EnDec RNN models, Sutskever~\textit{et al.}~\cite{sutskever2014sequence}, implemented a novel Sequence to Sequence (seq2seq) deep model using LSTM cells capable of taking arbitrary length input-output sequences. This model performs better than the RNN-based EnDec solutions, the authors advocate. 

Besides the above advanced recurrent architectures, there is a notion of adopting the standard 2D Convolutional Neural Networks (CNNs) for sequence data analysis. For instance, Bai~\textit{et al.}~\cite{bai2018empirical} introduced a practical solution, named Temporal Convolutional Network that applies convolutional operations by sliding the
1D-kernels over a specific length of temporal data. Similar to the canonical RNN, the TCN is a causal system such that a response at time stamp $t$ is the result of 1D-kernels convolved only with input quantities from time stamp $t$ and before in the earlier time stamps. Thus, the TCN can be viewed as 1D fully convolutional networks (FCN) with causal convolutions. It is found that the TCN outperforms generic RNN, GRU, and LSTM models in certain applications, like polyphonic music modeling, character/word-level language modeling, and IoT anomaly detection~\cite{bai2018empirical,9458520, wang2021lightlog}. 

\subsection{PMSM Temperature Prediction}

As discussed earlier, the temperature prediction is a time-series data analytical problem that can be handled through various modeling techniques. In this line, Wallscheid~\textit{et al.} the organizers of the \href{https://www.kaggle.com/wkirgsn/electric-motor-temperature}{PMSM benchmark dataset} from the Power Electrics and Drive Laboratory at Paderborn University introduced series of interesting models, ranging from conventional thermal equivalent circuit-based analysis---lumped-parameter thermal network (LPTN)~\cite{gedlu2020permanent} to modern deep learning-based models~\cite{wallscheid2017investigation, kirchgassner2020estimating}. 
For instance, the researchers in~\cite{wallscheid2017investigation} investigated predicting PMSM components' temperature with LSTM and GRU models. The authors found that predicting multiple targets value simultaneously was less accurate than predicting them separately. They explained the reason could be the different lagging characters of the target variables causing the target variable to have different dynamic responses to the input changes. The authors also suggested that applying data enrichment methods, like deriving new input variables from the raw input quantities by computing the absolute values of complex-valued inputs, statistical first moment (expected mean), and the second central moment (the variance) would aid the accurate prediction of the outputs. 

Similarly, Kirchgassner~\textit{et al.} in their two recent works \cite{kirchgassner2019deep} and \cite{kirchgassner2020estimating}, respectively in 2019 and 2020, applied TCN and residual LSTM for the task of PMSM's internal components' thermal stress estimation. Their studies show that the TCN model can generate more accurate results than the residual LSTM model. On the other hand, Lee~\textit{et al.}~\cite{9141245} proposed a Feed-forward Neural Network (FNN) based on Nonlinear Auto-Regressive Exogenous (NARX) to estimate the temperature of the permanent magnet and the stator winding in a PMSM. The authors indicate that their NARX structure would produce better results than the RNN and LSTM-based models if the relationship between the input and the output is known.

\section{Methodology}\label{methodoloy}

This section elaborates the data enrichment processes and the systematic implementation of various PMSM temperature predictive models. Fig.~\ref{fig_process_flow} depicts the steps involved in the proposed methodology and illustrates a deployment scenario of the proposed model.

\subsection{Data Preprocessing}

This work uses the benchmark dataset publicly provided by \href{https://www.kaggle.com/wkirgsn/electric-motor-temperature}{Power Electrics and Drive Laboratory} at the Paderborn University. 
It has 185-hour recordings of thirteen attributes of a PMSM. Among the thirteen attributes, twelve of them are continuous valued measurements and the last one is a discrete valued profile ID to distinguish sixty-nine real-world load scenarios of PMSMs (cf.~Table~\ref{tab:datadesc}). 
In this work, the 65th profile (profile\_id = 65) is reserved as test set following~\cite{kirchgassner2020estimating} for fair comparative analysis, while all other profiles are considered for training. 

\begin{table}[!t]
\renewcommand{\arraystretch}{1.3}
\caption{Attribute Description of the Dataset}
\label{tab:datadesc}
\centering
\begin{tabular}{|c||p{0.65\linewidth}|}
    \hline
    \textbf{Attribute Name} & \hspace{2cm} \textbf{Description} \\\hline
    \hline
    stator\_yoke & Stator yoke temperature (in °C) measured with thermocouples \\
    stator\_winding & Stator winding temperature (in °C) measured with thermocouples\\
    stator\_tooth & Stator tooth temperature (in °C) measured with thermocouples\\
    pm & Permanent magnet temperature (in °C) measured with thermocouples and transmitted wirelessly via a thermography unit.\\
    u\_q or $u_q$ & Voltage q-component measurement in dq-coordinates (in V)\\
    coolant & Coolant temperature (in °C)\\
    u\_d or $u_d$ & Voltage d-component measurement in dq-coordinates\\
    motor\_speed or $\omega$ & Speed of the PMSM (in rpm)\\
    i\_d or $i_d$ & Current d-component measurement in dq-coordinates\\
    i\_q or $i_q$ & Current q-component measurement in dq-coordinates\\
    ambient & Ambient temperature (in °C)\\
    torque & Motor torque (in Nm) \\
    profile\_id & Distinct measurement session identifier\\
    \hline \hline
\end{tabular}
\end{table}

\subsubsection{Attribute Selection and Synthetic Attribute Derivation}

\begin{figure*}[!ht]
\centering
\includegraphics[trim={0.2cm, 0.2cm, 4.7cm, 1.0cm}, clip, width=0.99\textwidth]{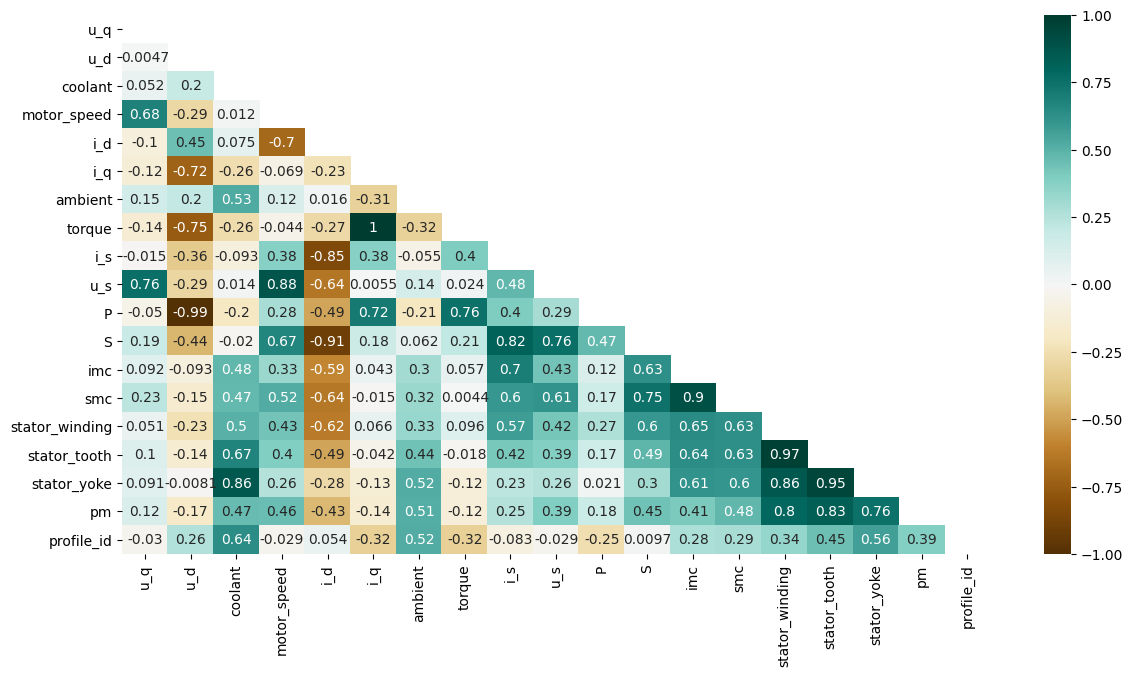}
\caption{Correlation between all the attributes of the PMSM generated using entire data samples.}
\vspace{-0.2cm}
\label{fig:correlation}
\end{figure*}

It is crucial to identify the attributes that are strong predictors of the target variables. In this work, the attributes \texttt{ambient, motor\_speed, coolant, u\_q, u\_d, i\_d,} and  \texttt{i\_q} are the predictor variables, while \texttt{stator\_yoke, stator\_winding, stator\_tooth,} and \texttt{pm} are the target variables.
To select the key attributes, we employ attribute correlation heat map as shown in Fig.~\ref{fig:correlation}. From this correlation map, it is found that the average absolute correlation (AvgAbsCor) of \texttt{torque} equals 0.089, which is the least compared to other input attributes' AvgAbsCor to the target attributes. Thus, \texttt{torque} is excluded from the input attribute list of the proposed models.

From the selected attributes, {six} synthetic features are constructed using linear and non-linear interaction of two input quantities as defined in~\eqref{norm-v} - \eqref{pwr1-speed} following the works ~\cite{wallscheid2017investigation, kirchgassner2019deep, kirchgassner2020estimating}. Through exhaustive empirical study this work, additionally, identifies two more useful interactions between coolant and current magnitude, and coolant and apparent power. These interactions are expressed in~\eqref{i-coolant} and \eqref{pwr1-coolant}, resulting in construction of two more synthetic attributes that boost the performances of the proposed models. While \eqref{i-coolant} shows the interaction between current and coolant temperature as a unified product quantity, \eqref{pwr1-coolant} express the interaction between the computed apparent power in \eqref{pwr1} and coolant temperature as a unified product quantity.
All these eight constructed attributes are further conditioned through Exponentially Weighted Moving Average (EWMA) operations elaborated in Section~\ref{sec:ewma} while training the proposed models. 


\begin{equation}\label{norm-v}
    Voltage~magnitude~(U) = \sqrt{u_d^2  + u_q^2}.
\end{equation}
\begin{equation}\label{norm-i}
    Current~magnitude~(I) = \sqrt{i_d^2  + i_q^2}.
\end{equation}
\begin{equation}\label{pwr1}
    Apparent~power~(S) = U \times I.
\end{equation}
\begin{equation}\label{pwr2}
    Effective~power~(P) = u_d \cdot i_d + u_q \cdot i_q. 
\end{equation}
\begin{equation}\label{i-speed}
    Current\text{-}motor\_speed~(IMM) = I \times \omega.
\end{equation}
\begin{equation}\label{pwr1-speed}
    Power\text{-}motor\_speed~(SMM) = S \times \omega.
\end{equation}
\begin{equation}\label{i-coolant}
    Current\text{-}coolant~(IMC) = I \times coolant.
\end{equation}
\begin{equation}\label{pwr1-coolant}
    Power\text{-}coolant~(SMC) = S \times coolant.
\end{equation}

\begin{figure*}[!ht]
\centering
\includegraphics[trim={3.9cm, 0.5cm, 3.7cm, 0.9cm}, clip, width=0.99\textwidth]{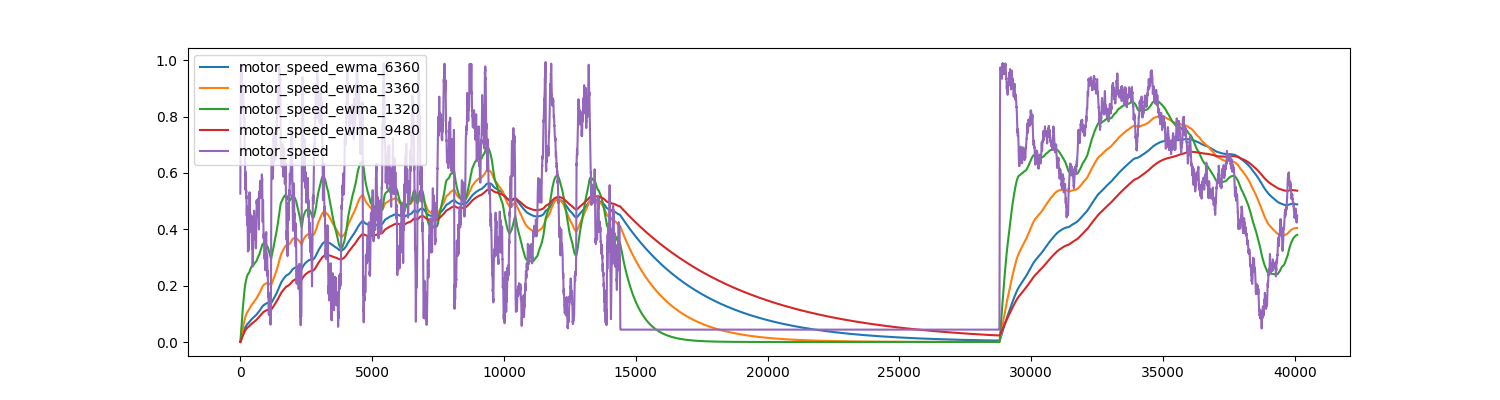}
\footnotesize{ Sample ID}
\caption{Visualizing EWMA transformation with span values of 1320, 3360, 6360, and 9480 of \textcolor{Fuchsia}{motor speed}.}
\label{fig:ewma_motor_speed}
\end{figure*}
\begin{figure*}[!ht]
\centering
\includegraphics[trim={3.9cm, 0.5cm, 3.7cm, 0.9cm}, clip, width=0.99\textwidth]{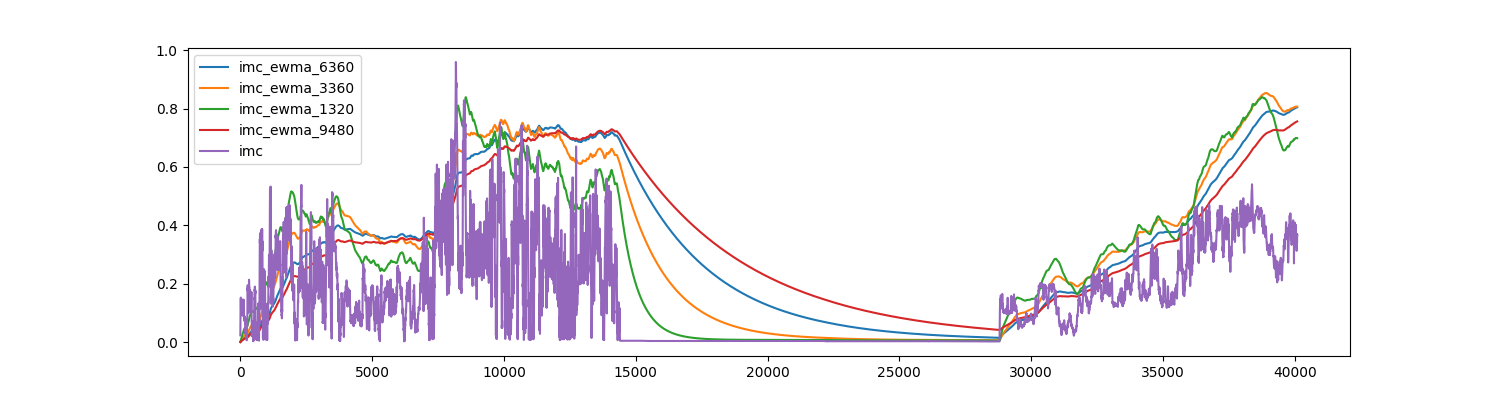}
\footnotesize{ Sample ID}
\caption{Visualizing EWMA transformation with the span values of 1320, 3360, 6360, and 9480 of the synthetically constructed feature, \textcolor{Fuchsia}{IMC} (cf.~\eqref{i-coolant}).} 
\label{fig:ewma_imc}
\end{figure*}

\subsubsection{Exponentially Weighted Moving Average}\label{sec:ewma}

The raw measurements of the PMSM have high fluctuation, noise, and the target measures have inconsistent amount of lag among them. 
For example, the permanent magnet temperature has greater lag than other three target variables, since the permanent magnet is placed in the core (cf. component\#5 in ~Fig.~\ref{fig:pmsm}) and it is heated and cooled passively. It shows a property of low-pass filter at heat dissipation. Therefore, this work apply EWMA as a data enrichment process to counter the impact of the aforesaid variations in the raw data streams. 
EWMA defined in~\eqref{eq_ewma} has been an effective data preprocessing method for PSPM temperature prediction~\cite{kirchgassner2019empirical}. 
\begin{equation} \label{eq_ewma}
    y_t=\frac{x_t+(1-\alpha)x_{t-1}+(1-\alpha)^2x_{t-2}+\cdots+(1-\alpha)^{t}x_0}{1+(1-\alpha)+(1-\alpha)^2+\cdots+(1-\alpha)^t},
\end{equation} 
where $y_t$ is the output of transforming an input sequence, $\{x_0, x_1, \cdots, x_t\}$, and $(1-\alpha)$ is the weight. By letting $\omega_i = (1-\alpha)^i$, \eqref{eq_ewma} can condensed in a closed-form summation expression as
\begin{equation}\label{eq_ewma_sum}
    y_t = \frac{\sum_{i=0}^t \omega_ix_{t-i}}{\sum_{i=0}^t \omega_i},
\end{equation}
where $\alpha = 2/(s+1)$ with $s$ is being the \textit{span} or the \textit{lookback period} defined by the user. The weight, $\omega_i$ decides how significant the observation, $x_{t-i}$ is in the computation of EWMA, such that weights decrease exponentially as the observation gets older. 
The span, in this work, is the number of samples to be considered when computing $y_t$. The shorter the span the closer the EWMA tracks the raw time series data, while a larger span value produces more smoothed version of the input. Examples of EWMA transformation of the attributes using~\eqref{eq_ewma} are shown in Fig.~\ref{fig:ewma_motor_speed} and Fig.~\ref{fig:ewma_imc} with various span values using Pandas \texttt{ewm()} function\footnote[2]{\url{https://pandas.pydata.org/docs/reference/api/pandas.Series.ewm.html?highlight=ewma}}. Fig.~\ref{fig:ewma_motor_speed} visualizes EWMA transformation of raw motor speed quantities, and Fig.~\ref{fig:ewma_imc} visualizes EWMA transformation of the synthetically constructed attribute using \eqref{i-coolant}, where one can find that the EWMA has produced more smoothed version of the motor speed for $s=9480$ compared to all other span values.

\subsection{Models}\label{proposed-models}

\setlength{\tabcolsep}{2pt}
\begin{table}[!t]  \caption{Layer Connectivity Detail of the LSTM Encoder-Decoder Model.\\ $\beta$ represents the batch size.}
    \centering
    \begin{tabular}{|p{1.8cm}|p{1.5cm}|p{2.5cm}|p{2.5cm}|}
    \hline \hline
     \textbf{Layer}    &  \textbf{Type} & \textbf{Shape} & \textbf{Connect To} \\ \hline \hline
     Input & Input & ($\beta$, 180, 65) & \\ 
     & Output & ($\beta$, 180, 65) & Encoder Input\\ \hline 
    Encoder & Input & ($\beta$, 180, 65) & \\
    & Output-1 & ($\beta$, 100) & RepeatVector Input\\
    & Output-2 & [($\beta$, 100), ($\beta$, 100)] & Decoder Input-2\\ \hline  
    RepeatVector & Input & ($\beta$, 100) & \\
    & Output & ($\beta$, 1, 100) & Decoder Input-1\\ \hline
    Decoder & Input-1 & ($\beta$, 1, 100) & \\
    & Input-2 & [($\beta$, 100), ($\beta$, 100)] & RepeatVector Input\\
    & Output & ($\beta$,1, 100) & Dense Input\\ \hline
    Dense & Input & ($\beta$, 1, 100) & \\ 
     & Output & ($\beta$, 1, 4) & \\ \hline
     
     \multicolumn{4}{|c|}{Total trainable parameters: 147,204} \\\hline \hline
    \end{tabular}
    \label{tab:lstm-endec}
    \vspace{-0.2cm}
\end{table}

\setlength{\tabcolsep}{2pt}
\begin{table}[!t]  \caption{Layer Connectivity Detail of the BiLSTM Encoder-Decoder Model.\\ $\beta$ represents the batch size.}
    \centering
    \begin{tabular}{|p{1.8cm}|p{1.25cm}|p{2.4cm}|p{2.9cm}|}
    \hline \hline
     \textbf{Layer}    &  \textbf{Type} & \textbf{Shape} & \textbf{Connect To} \\ \hline \hline
     Input & Input & ($\beta$, 180, 65) & \\ 
     & Output & ($\beta$, 180, 65) & Encoder Input\\ \hline 
    Encoder & Input & ($\beta$, 180, 65) & \\
    & Output-1 & ($\beta$, 100) & Concat-1 Input\\
    & Output-2 & [($\beta$, 100), ($\beta$, 100)] & Concat-2 Input\\ \hline  
    Concat-1 & Input & [($\beta$, 100), ($\beta$, 100)] & \\
    & Output & ($\beta$, 200) & RepeatVector Input, Decoder Input-2\\ \hline
    Concat-2 & Input & [($\beta$, 100), ($\beta$, 100)] & \\
    & Output & ($\beta$, 200) & Decoder Input-3\\ \hline
    RepeatVector & Input & ($\beta$, 200) & \\
    & Output & ($\beta$, 1, 200) & Decoder Input-1\\ \hline
    Decoder & Input-1 & ($\beta$, 1, 200) & \\
    & Input-2 & ($\beta$, 200) & \\
    & Input-3 & ($\beta$, 200) & \\
    & Output & ($\beta$,1, 200) & Dense Input\\ \hline
    Dense & Input & ($\beta$, 1, 200) & \\ 
     & Output & ($\beta$, 1, 4) & \\ \hline \multicolumn{4}{|c|}{Total trainable parameters: 454,404} \\\hline \hline 
    \end{tabular}
    \label{tab:bidir-lstm-endec}
    \vspace{-0.5cm}
\end{table}

\setlength{\tabcolsep}{2pt}
\begin{table}[!ht]  \caption{Layer Connectivity Detail of the Attention-based LSTM Encoder-Decoder Model. $\beta$ represents the batch size.}
    \centering
    \begin{tabular}{|p{1.8cm}|p{1.25cm}|p{2.4cm}|p{2.9cm}|}
    \hline \hline
     \textbf{Layer}    &  \textbf{Type} & \textbf{Shape} & \textbf{Connect To} \\ \hline \hline
     Input & Input & ($\beta$, 180, 65) & \\ 
     & Output & ($\beta$, 180, 65) & Encoder Input\\ \hline 
    Encoder & Input & ($\beta$, 180, 65) & \\
    & Output-1 & ($\beta$, 100) & RepeatVec. Input\\
    & Output-2 & [($\beta$, 100), ($\beta$, 100)] & Decoder Input-2\\ 
    & Output-3 & ($\beta$, 180, 100) & Dot-1 Input-2, Dot-2 Input-2\\\hline  
  
    RepeatVec. & Input & ($\beta$, 100) & \\
    & Output & ($\beta$, 1, 100) & Decoder Input-1\\ \hline
    Decoder & Input-1 & ($\beta$, 1, 100) & \\
    & Input-2 & [($\beta$, 100), ($\beta$, 100)] & \\
    & Output & ($\beta$, 1, 100) & Dot-1 Input-1, Concat Input-1\\ \hline
    Dot-1 & Input-1 & ($\beta$, 1, 100) & \\
    & Input-2 & ($\beta$, 180, 100) & \\
    & Output & ($\beta$, 1, 180) & Dot-2 Input-1 \\ \hline
    Dot-2 & Input-1 & ($\beta$, 1, 180) & \\
    & Input-2 & ($\beta$, 180, 100) & \\
    & Output & ($\beta$, 1, 100) & Concat Input-2 \\ \hline
    Concat & Input-1 & ($\beta$, 1, 100) & \\
    & Input-2 & ($\beta$, 1, 100) & \\
    & Output & ($\beta$, 1, 200) & Dense Input\\ \hline
    Dense & Input & ($\beta$, 1, 200) & \\ 
     & Output & ($\beta$, 1, 4) & \\ \hline \multicolumn{4}{|c|}{Total trainable parameters: 147,604} \\\hline \hline 
    \end{tabular}
    \label{tab:atten-lstm-endec}
\end{table}

\begin{figure}[!t]	
	\includegraphics[trim={0.8cm, 3.2cm, 0.6cm, 2.8cm}, clip,width=0.95\columnwidth]{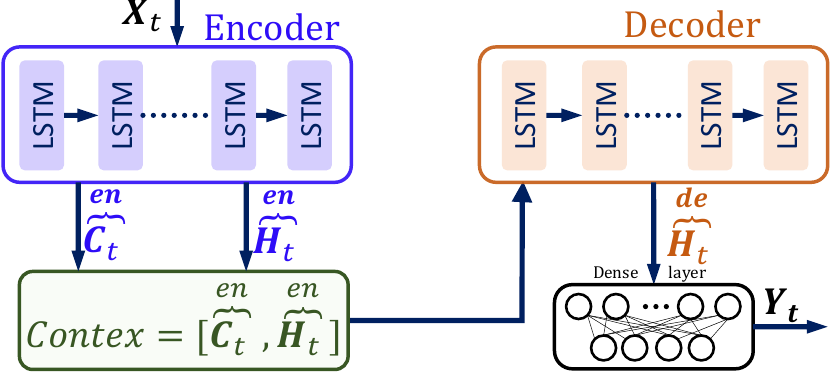}
	\caption{Illustration of the vanilla EnDec LSTM: $\mathbf{X}_t$, $\LaTeXoverbrace {H_{t}}^{en}$, $\LaTeXoverbrace{C_{t}}^{en}$, $\LaTeXoverbrace{H_t}^{de}$, and $\mathbf{Y}_t$ stand for the input sequence, encoder's hidden sequence, encoder's cell state, decoder hidden sequence, and the output sequence, respectively, at time $t$.}
	\label{fig:lstm_ende_cell}
\end{figure}

As discussed in Section~\ref{seq-modeling}, the PMSM's temperature prediction is a time-series regression problem. It can be effectively modeled through EnDec DNN architectures as used in several Natural Language Processing (NLP) tasks, for instance, language translation. 
Therefore, this work exploits the EnDec structure as a backbone to build three models with different configurations of LSTM blocks. The architectural details of the three models are as summarized in Tables~\ref{tab:lstm-endec}, \ref{tab:bidir-lstm-endec}, and \ref{tab:atten-lstm-endec}, while Fig.~\ref{fig:lstm_ende_cell} - Fig.~\ref{fig:attention-lstm_cell} illustrate their network structures. The first configuration is a vanilla EnDec LSTM model (cf.~Table~\ref{tab:lstm-endec}). The second configuration is a bidirectional EnDec LSTM model (cf.~Table~\ref{tab:bidir-lstm-endec}), while the third configuration improves the first model with a global attention mechanism (cf.~Table~\ref{tab:atten-lstm-endec}).
The existing studies show that building single target-specific models can provide better accuracy~\cite{guo2020predicting, 9141245}. However, it becomes burdensome to have multiple models. Thus, this work endeavors to come up with unified models that can predict all target quantities, viz. the temperatures of stator yoke, stator winding, stator tooth and permanent magnet, simultaneously. With this in mind, all three proposed models have a four-way output layer (cf.~Tables~\ref{tab:lstm-endec}, \ref{tab:bidir-lstm-endec}, and \ref{tab:atten-lstm-endec}).

\begin{figure}[!t]	
	\centering
\includegraphics[trim={4.5cm, 0.8cm, 7.0cm, 3.2cm}, clip,width=0.8\columnwidth]{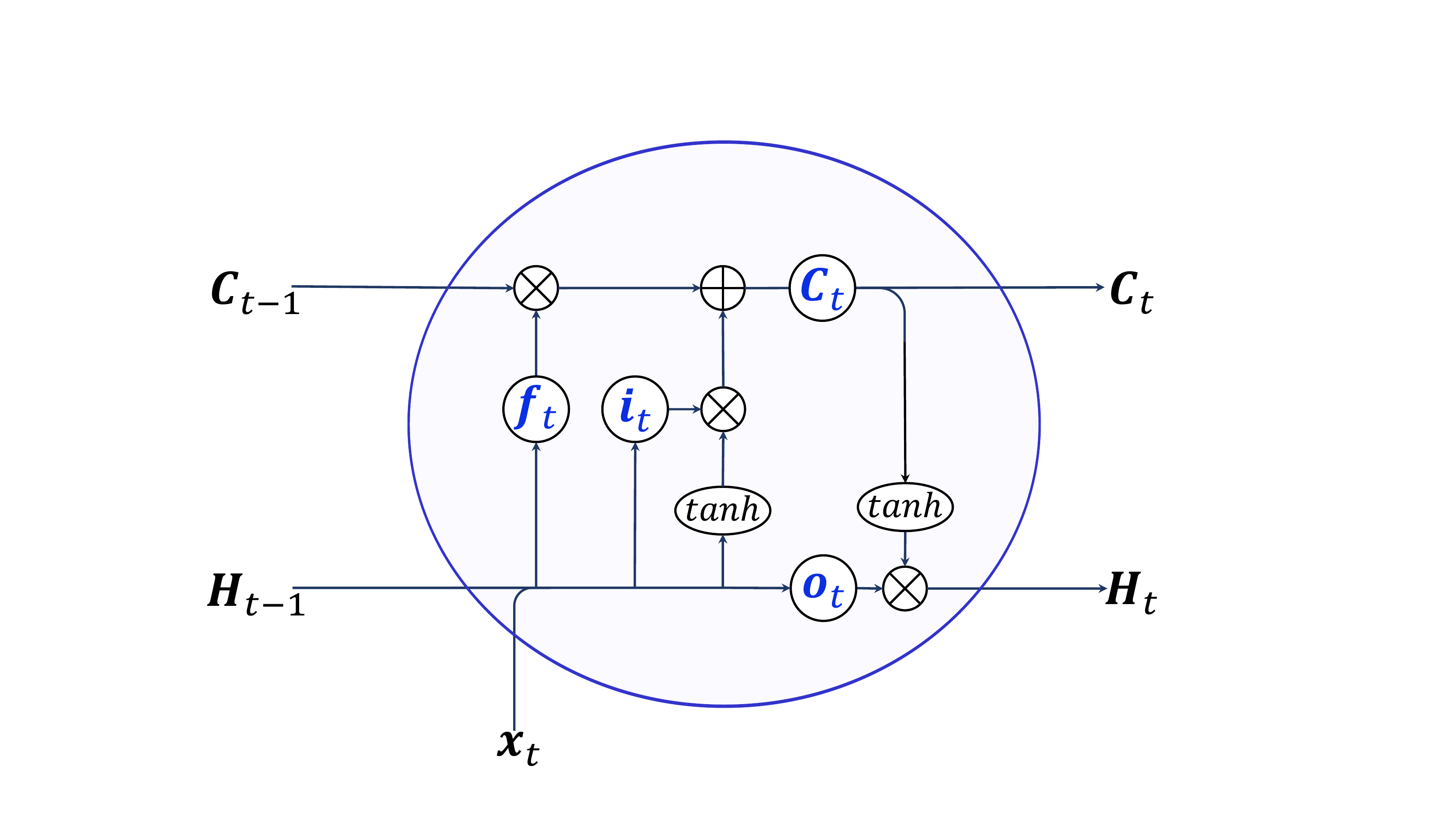}
	\caption{Illustration of a standard LSTM cell with three gates that control information flow from a time-series data, where $\mathbf{X}_t$, $\mathbf{C}_t$, $\mathbf{H}_t$, $\mathbf{i}_t$, $\mathbf{f}_t$ and $\mathbf{o}_t$ are the input quantity from a time-series data, cell state, hidden state, input gate, forget gate, and output gate, respectively, at timestamp, $t$.}
	\label{fig:lstm_cell}
\end{figure}

The LSTM units are well-known for their long-term information retention capability~\cite{akilan2019video, 8671459} due to their unique gated structure as shown in Fig.~\ref{fig:lstm_cell} and defined by \eqref{eqn_lstm-input} - \eqref{eqn_lstm-hidden}, where $X_t$ is an input quantity from a time-series data, $C_t$ is the cell state, $H_t$ is the hidden state, and $i_t$, $f_t$, and $o_t$ are the gates of the LSTM block at timestamp $t$. Hence, $W$, '\textasteriskcentered', and '$\circ$' denote conv kernels specific to the gates and internal states, the conv operator, and Hadamard product. The $\sigma$ is a $hard$ $sigmoid$ function.

\begin{gather} \label{eqn_lstm-input}
i_t = \sigma(W_{xi} \ast X_t + W_{hi} \ast H_{t-1} + b_i),\\
f_t = \sigma(W_{xf} \ast X_t + W_{hf} \ast H_{t-1} +b_f),\\
o_t = \sigma(W_{xo} \ast X_t + W_{ho} \ast H_{t-1} +b_o),\\
C_t = f_t \circ C_{t-1} + i_t \circ \tanh(W_{xc} \ast X_t + W_{hc} \ast H_{t-1} +b_c),\\
H_t = o_t \circ \tanh(C_{t}).\label{eqn_lstm-hidden}
\end{gather}

\subsection{{Model 1: Encoder-Decoder LSTM}}\label{sec:model1}
Model 1 is a vanilla encoder-decoder LSTM structure as illustrated in Fig.~\ref{fig:lstm_ende_cell} (cf.~Table~\ref{tab:lstm-endec}). The input sequence passes through the encoder and generates corresponding cell state and hidden state at time, $t$. Then, the decoder uses the concatenated vector of these two states of the encoder to updates its hidden state. Finally, the dense layers at the top of the network map the sequence of decoder hidden state to output, $\mathbf{Y}_t$ with a learned weight, $\mathbf{W}_y$ as defined by \eqref{eq_vallina_endec_lstm_yt}.
\begin{align} \label{eq_vallina_endec_lstm_yt}
  \mathbf{Y}_t = \mathbf{W}_y(\LaTeXoverbrace{H_t}^{de}). 
\end{align}
  
\subsection{{Model 2: Bidirectional Encoder-Decoder LSTM}}\label{sec:model2}
\begin{figure}[!t]	
	\includegraphics[trim={0.8cm, 2.5cm, 0.9cm, 2.6cm}, clip,width=0.95\columnwidth]{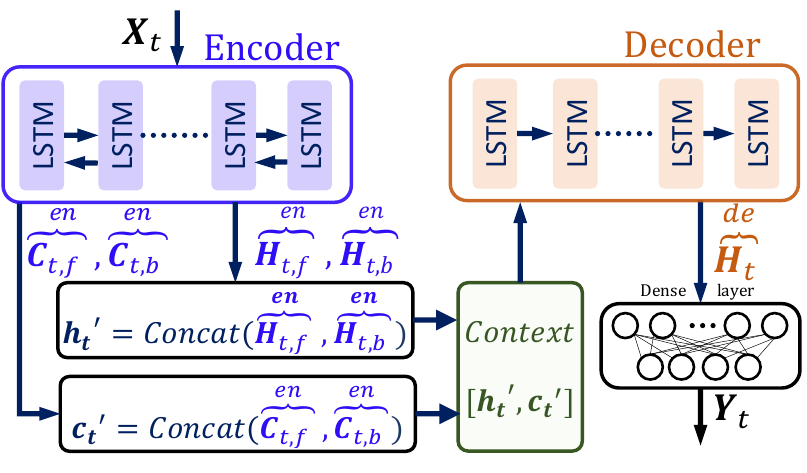}
	\caption{Illustration of the bidirectional EnDec LSTM: $\mathbf{X}_t$, $\LaTeXoverbrace{H_{t, f}}^{en}$, $\LaTeXoverbrace{C_{t, f}}^{en}$, $\LaTeXoverbrace{H_{t, b}}^{en}$, $\LaTeXoverbrace{C_{t, b}}^{en}$, $\LaTeXoverbrace{H_t}^{de}$, and $\mathbf{Y}_t$ stand for the input sequence, encoder's forward hidden sequence,  encoder's forward cell state, encoder's backward hidden sequence,  encoder's backward cell state, decoder hidden sequence, and the output sequence, respectively, at time $t$.}
	\label{fig:bidirectional-lstm_cell}
\end{figure}

Contrast to the vanilla EnDec LSTM model, in  bidirectional EnDec LSTM model, the dimension of the hidden and cell states are doubled due to the backward status updates, where the forward and backward status are concatenated to form a context vector as illustrated in Fig.~\ref{fig:bidirectional-lstm_cell}. Then, the decoder uses the context vector to updates its hidden state. Finally, the dense layers at the top of the network map the sequence of decoder hidden state to output, with a learned weight, $\mathbf{W}_y$ as defined by \eqref{eq_vallina_endec_lstm_yt} similar to the vanilla model.

\subsection{{Model 3: Global Attention-based EnDec LSTM}}\label{sec:model3}

\begin{figure}[!t]	
	\includegraphics[trim={0.35cm, 1.8cm, 0.8cm, 2.6cm}, clip,width=0.95\columnwidth]{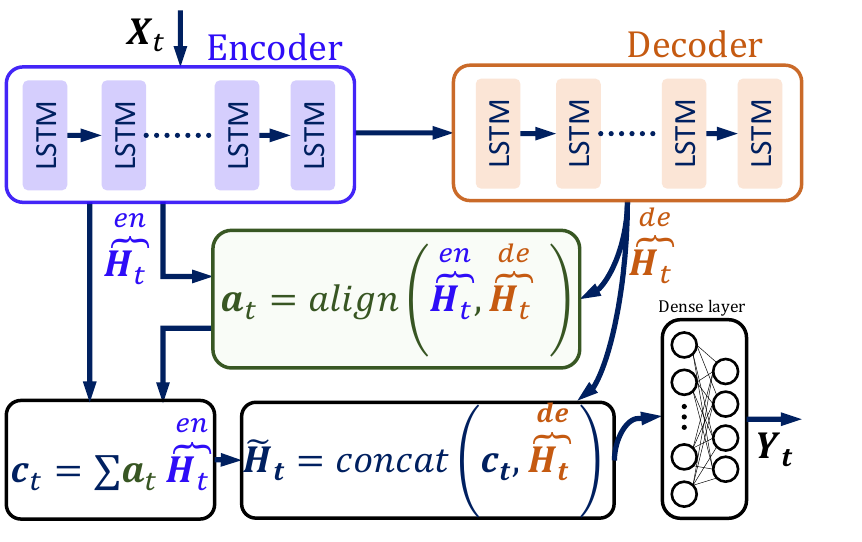}
	\caption{Illustration of the attention-based EnDec LSTM: $\mathbf{X}_t$, $\LaTeXoverbrace{H_t}^{en}$, $\LaTeXoverbrace{H_t}^{de}$, ${\widetilde{H}}_t$, $\mathbf{a}_t$, $\mathbf{c}_t$, and $\mathbf{Y}_t$ stand for the input sequence, encoder hidden sequence, decoder hidden sequence, attentional hidden state, attention vector, context vector, and output sequence, respectively, at time $t$.}
	\label{fig:attention-lstm_cell}
\end{figure}

In the global attention-based EnDec LSTM model (cf.~Fig.~\ref{fig:attention-lstm_cell}), the encoder hidden status sequence and the decoder hidden state sequence form an alignment wight vector, $\mathbf{a}_t$ through the computations defined in \eqref{eq_align} - \eqref{eq_score}. Then, this alignment vector is used to generate a global context vector, $\mathbf{c}_t$ that is computed as a weighted average of the entire encoder hidden sequence using \eqref{eq_ct}. The resulting context vector and the decoder's hidden status sequence are concatenated to generate the attentional hidden state, $\widetilde{H}_t$ as expressed in~\eqref{eq_atten_ht}. 
Eventually, it is then passed through dense layer(s) and conditioned with a linear activation having learned weight, $\mathbf{W}_y$ 
to compute the output, $\mathbf{Y}_t$ as defined in~\eqref{eq_yt}. 

\begin{align} \label{eq_align}
    \mathbf{a}_t &= align\left(\LaTeXoverbrace{H_t}^{en}, \LaTeXoverbrace{H_t}^{de}\right)\\
     & = \frac{exp\left(score(\LaTeXoverbrace{H_t}^{en}, \LaTeXoverbrace{H_t}^{de})\right)}{\sum{exp\left(score(\LaTeXoverbrace{H_t}^{en}, \LaTeXoverbrace{H_t}^{de})\right)}}, 
\end{align}
where $\LaTeXoverbrace{H_t}^{en}$, and $\LaTeXoverbrace{H_t}^{de}$ are the encoder hidden sequence, and decoder hidden sequence, respectively, at time $t$. And the $score()$ is computed as a dot product between the two hidden sequences:
\begin{align} \label{eq_score}
   score(\LaTeXoverbrace{H_t}^{en}, \LaTeXoverbrace{H_t}^{de}) = {\LaTeXoverbrace{H_t}^{en}}^T{\LaTeXoverbrace{H_t}^{de}}. 
\end{align}
\begin{align} \label{eq_ct}
  \mathbf{c}_t = \sum{\mathbf{a}_t \LaTeXoverbrace{H_t}^{en}}. 
\end{align}
\begin{align} \label{eq_atten_ht}
    \widetilde{H}_t & = concat(\mathbf{c}_t,  \LaTeXoverbrace{H_t}^{de}) 
\end{align}
\begin{align} \label{eq_yt}
  \mathbf{Y}_t = \mathbf{W}y(\widetilde{H}_t) . 
\end{align}

\begin{table*}[!t] 
\caption{Performance Comparison of Various Models wrt MSE and MAE. The Best Performance is Highlighted in \textcolor{ForestGreen}{Green Ink} and \textbf{Boldface Text}. \\ \footnotesize{Note: Smallest MSE Value Indicates the Best Performance} and "-" Indicates The Quantity Is Not Available in the Literature.} \label{tab:results}
\setlength\tabcolsep{6pt}
    \centering
    \begin{tabular}{|c|c|c|c|c|c|c|c|}
        \hline \hline
        \multirow{2}[1]{*}{\textbf{Targets}}  & \multirow{2}[1]{*}{\textbf{Metrics}} & \multicolumn{3}{c}{\textbf{Other Models}}  & \multicolumn{3}{|c|}{\textbf{Our Models}}\\ \cline{3-5} \cline{6-8} 
        & & TCN~\cite{kirchgassner2020estimating} & LTPN~\cite{gedlu2020permanent}& RNN~\cite{kirchgassner2020estimating} & Model 1 & Model 2 & Model 3 \\ \hline \hline
        \multirow{2}[1]{*}{{Stator Winding}}  & MSE & 6.90 & - & - & 4.86 & 5.28 & 2.82 \\ 
                                                     & MAE & 7.92 & - & & 8.47 & 10.09 & 8.75 \\ \hline
       \multirow{2}[1]{*}{{Stator Tooth }}    & MSE & 2.84 & - & - & 2.23 & 2.82 & 1.84 \\ 
                                                     & MAE & 6.24 & - & & 4.89 & 4.89 & 6.05 \\ \hline
       \multirow{2}[1]{*}{{Stator Yoke}}      & MSE & 1.80 & - & - & 1.34 & 2.04 & 1.04 \\ 
                                                     & MAE & 5.24 & - & - & 4.03 & 4.58 & 3.51 \\ \hline
       \multirow{2}[1]{*}{{Permanent Magnet}} & MSE & 0.65 & - & - & 1.52 & 5.61 & 1.17 \\ 
                                                     & MAE & 5.84 & - & - & 5.62 & 6.0 & 3.05 \\ \hline
       \multirow{2}[1]{*}{\textbf{Overall}}         & MSE & 3.04 & $5.73^*$ & $8.70$ & 2.49 & 3.94 & \textcolor{ForestGreen}{\textbf{1.72}} \\ 
                                                     & MAE & 6.31 & - & - & 5.75 & 6.39 & \textcolor{ForestGreen}{\textbf{5.34}}\\ \hline
        \multicolumn{5}{|c}{Inference time per batch ($ms$)}  & $\approx$ 13 & $\approx$ 19 & $\approx$ 13\\
        
         \hline \hline
         \multicolumn{8}{l}{*as reported in \cite{kirchgassner2020estimating}}

    \end{tabular}

\end{table*}

\section{Experimental Results and Discussion}\label{experimental-results}

\subsection{Experimental Setup}\label{data-preprocessing}

\begin{table}[!t]
\caption{Summary of Hyperparameter Setting}
    \centering
    \begin{tabular}{|l|l|}
    \hline \hline
    \textbf{Hyperparameter} & \textbf{Value is Set to: } \\\hline\hline
    Loss function & MSE (cf.~\eqref{eq_mse})\\
    Input sequence length & 180\\
    Hidden dimension & 100\\
    Output dimension & $1\times4$\\
    Batch size ($\beta$) & 256\\
    Optimizer & Adam\\
    Learning rate & 0.0005\\
    Hidden layer activation & Tanh\\
    \hline\hline
\end{tabular}
\label{tab:hyperpra}
\end{table}

All the models described in Tables~\ref{tab:lstm-endec}, \ref{tab:bidir-lstm-endec}, and \ref{tab:atten-lstm-endec} share the same hyperparameters tabulated in Table~\ref{tab:hyperpra}. Additionally, model-3 uses a global attention mechanism as introduced in \cite{luong2015effective}.

\subsection{Computing Platform}
The hardware platform is a laptop having an AMD 5800H CPU (3.2GHz), a RTX 3060 GPU and 32GB RAM. Meanwhile, Windows 10 and Tensorflow 2.6.0 are used as the software platform.\\

\subsection{Evaluation Metrics}

Since this work handles a regression problems (i.e., real value prediction), the mean squared error (MSE) defined in \eqref{eq_mse} is used as the primary evaluation metric. The smaller MSE value indicates better performance of the model under study.
\begin{equation}
    MSE = \dfrac{1}{n} \sum_{i=1}^n(y_i - \hat{y}_i)^2, i={1, 2, \cdots n},
    \label{eq_mse}
\end{equation}
where $n$, $y_i$, $\hat{y}_i$, and $i$ are the total number of data points predicted, true value, the predicted value, and sample index, respectively. 
For more comparative study with existing works, the maximum absolute error (MAE) is adopted as the secondary assessment indicator. It can show the maximum deviation from the ground truth as expressed by~\eqref{eq_mae}.
\begin{equation}
    MAE = max(|y_i - \hat{y}_i|), i={1, 2, \cdots n}.  
    \label{eq_mae}
\end{equation}

\subsection{Training strategy}
Due to the small amount of the GPU memory, the training dataset can not be accessed as a whole. The alternative approach includes two steps. The first step divides the whole dataset into four groups and training each group one by one. In the next step, some randomly selected load profiles among the training data are used to form a new group to fine tuned the models to reach the final generalized models.

\begin{figure*}[!t]
  \centering 
    \begin{tabular}{p{4.2cm}p{4.2cm}p{4.2cm}p{4.2cm}}
        \centering \footnotesize{Stator Winding} & \centering \footnotesize{Stator Tooth} & \centering \footnotesize{Stator Yoke} & \centering \footnotesize{Permanent Magnet} \\
    \end{tabular}
    \vspace{-0.5cm}
    
\begin{center}
     {
    \includegraphics[trim={0.3cm, 0.4cm, 0.4cm, 0.2cm}, clip, width=1.0\textwidth]{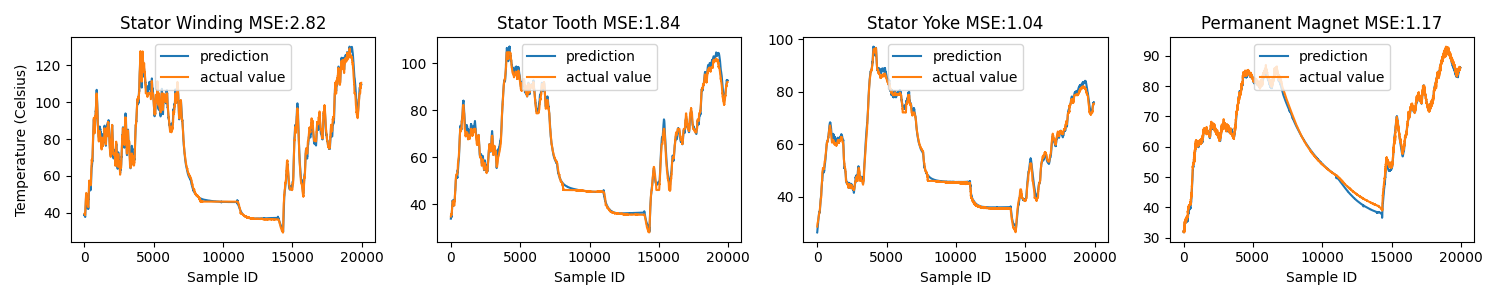}
    
    \vspace{0.1cm}
    \footnotesize{Prediction vs Actual: x, y axes represent test set sample IDs and the temperature readings in °C.}
    }
    
    \begin{tabular}{p{4.2cm}p{4.2cm}p{4.2cm}p{4.2cm}}
        \centering \footnotesize{Stator Winding} & \centering \footnotesize{Stator Tooth} & \centering \footnotesize{Stator Yoke} & \centering \footnotesize{Permanent Magnet} \\
    \end{tabular}
    \vspace{-0.3cm}
    
   { \includegraphics[trim={0cm, 0.4cm, 0.4cm, 0.2cm}, clip, width=1.0\textwidth]{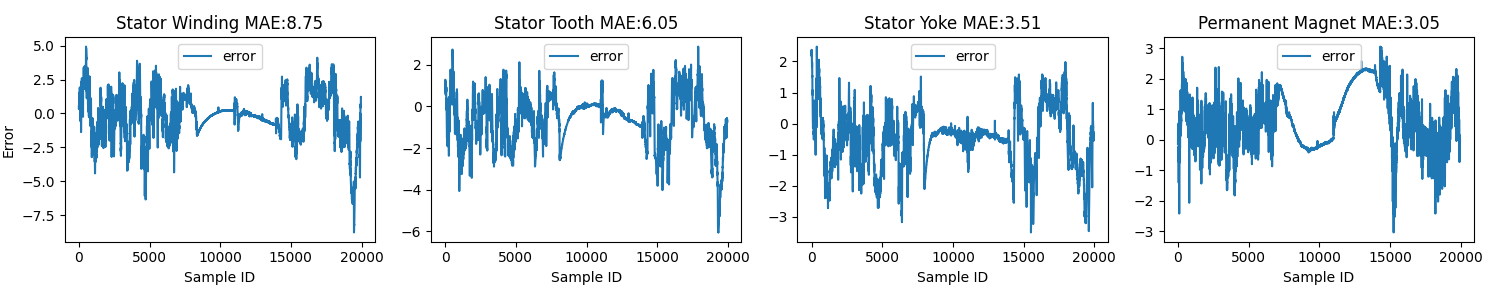}}   \\
    \vspace{0.1cm}
    \footnotesize{The computed error term between the predicted and actual temperature readings: x, y axes represent test set sample IDs and error values.}
    
 \end{center}
    
\caption{Performance evaluation of the proposed attention-based encoder-decoder LSTM architecture (cf.~Model 3 in Section~\ref{sec:model3}).}
 \label{fig:mse_mae_plot_atten_endec_lstm}
   
\end{figure*}

\begin{figure*}[!t]
  \centering 
    \begin{tabular}{p{4.2cm}p{4.2cm}p{4.2cm}p{4.2cm}}
        \centering \footnotesize{Stator Winding} & \centering \footnotesize{Stator Tooth} & \centering \footnotesize{Stator Yoke} & \centering \footnotesize{Permanent Magnet} \\
    \end{tabular}
    \vspace{-0.5cm}
    
\begin{center}
     { 
    \includegraphics[trim={0.3cm, 0.4cm, 0.4cm, 0.2cm}, clip, width=1.0\textwidth]{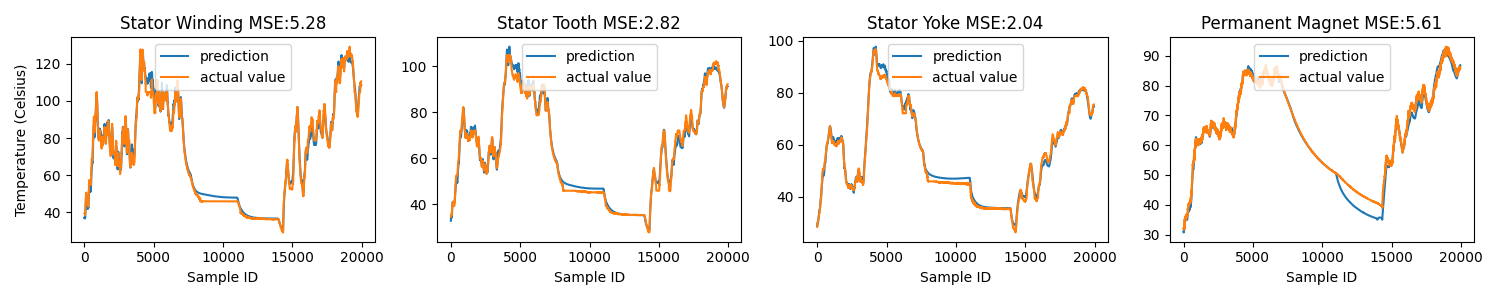}
    \vspace{0.1cm}
    \footnotesize{Prediction vs Actual: x, y axes represent test set sample IDs and the temperature readings in °C.}
    }
    
    \begin{tabular}{p{4.2cm}p{4.2cm}p{4.2cm}p{4.2cm}}
        \centering \footnotesize{Stator Winding} & \centering \footnotesize{Stator Tooth} & \centering \footnotesize{Stator Yoke} & \centering \footnotesize{Permanent Magnet} \\
    \end{tabular}
    \vspace{-0.3cm}

   \includegraphics[trim={0.3cm, 0.4cm, 0.4cm, 0.2cm}, clip, width=1.0\textwidth]{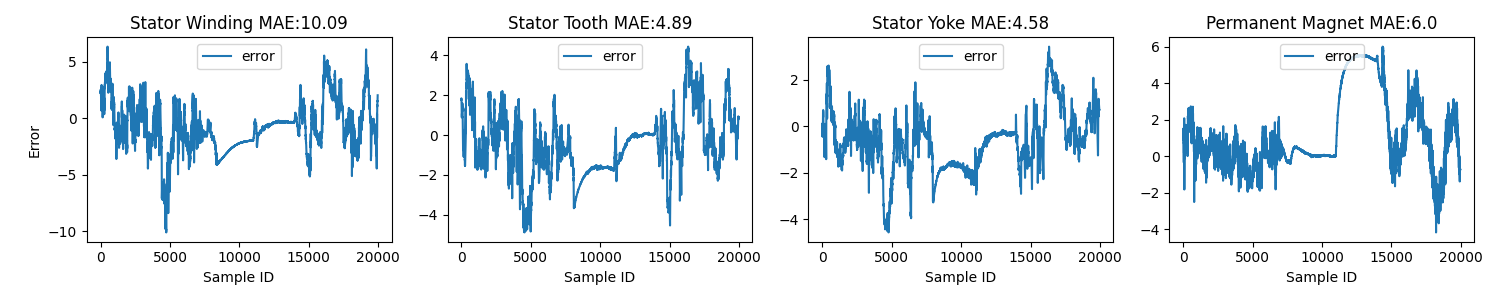}
    \footnotesize{The computed error term between the predicted and actual temperature readings: x, y axes represent test set sample IDs and error values.}
    
 \end{center}
    
\caption{Performance evaluation of the proposed Bidirectional LSTM encoder-decoder architecture (cf.~Model 2 in Section~\ref{sec:model2}).}
 \label{fig:mse_mae_plot_bidir_endec_lstm}
   
\end{figure*}

\subsection{Overall Analysis}

\begin{figure*}[!t]
  \centering 
    \begin{tabular}{p{4.2cm}p{4.2cm}p{4.2cm}p{4.2cm}}
        \centering \footnotesize{Stator Winding} & \centering \footnotesize{Stator Tooth} & \centering \footnotesize{Stator Yoke} & \centering \footnotesize{Permanent Magnet} \\
    \end{tabular}
    \vspace{-0.5cm}
    
\begin{center}
     {
    \includegraphics[trim={0.3cm, 0.4cm, 0.4cm, 0.2cm}, clip, width=1.0\textwidth]{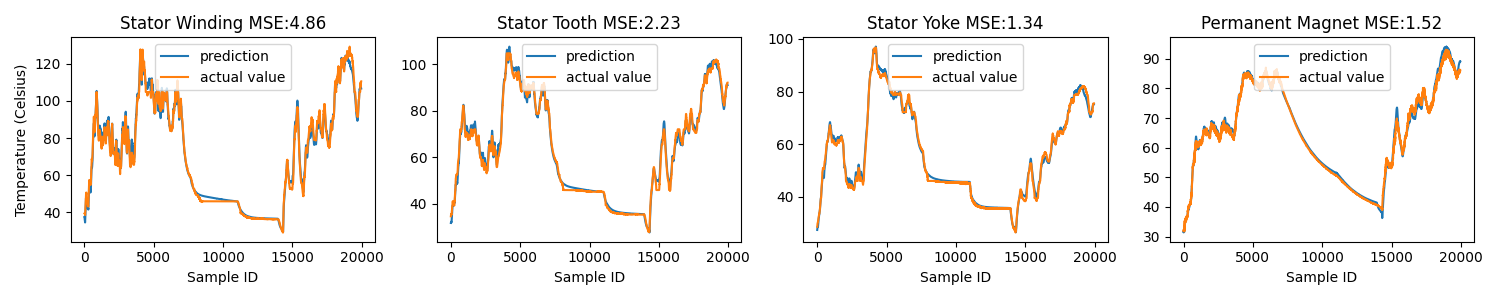}
    \vspace{0.1cm}
    \footnotesize{Prediction vs Actual: x, y axes represent test set sample IDs and the temperature readings in °C.}
    }
    
    \begin{tabular}{p{4.2cm}p{4.2cm}p{4.2cm}p{4.2cm}}
        \centering \footnotesize{Stator Winding} & \centering \footnotesize{Stator Tooth} & \centering \footnotesize{Stator Yoke} & \centering \footnotesize{Permanent Magnet} \\
    \end{tabular}
    \vspace{-0.3cm}

   \includegraphics[trim={0.0cm, 0.4cm, 0.4cm, 0.2cm}, clip, width=1.0\textwidth]{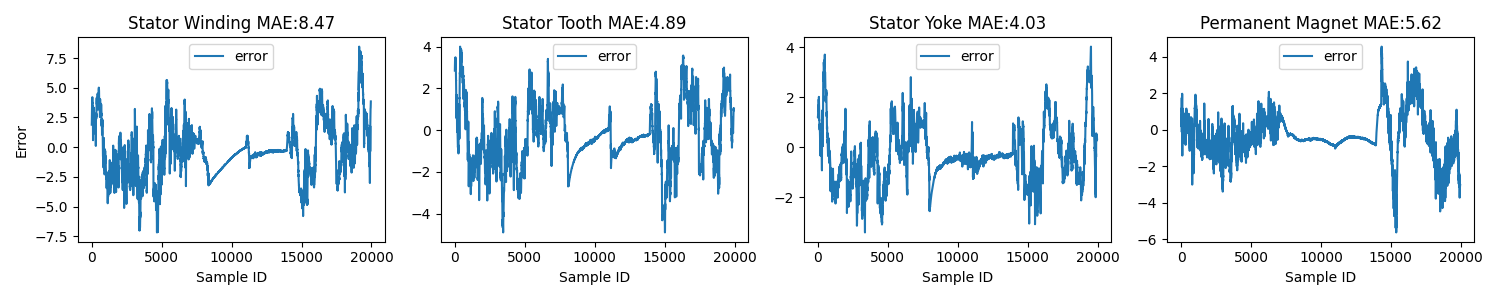}
    \footnotesize{The computed error term between the predicted and actual temperature readings: x, y axes represent test set sample IDs and error values.}
    
 \end{center}
    
\caption{Performance evaluation of the proposed LSTM encoder-decoder architecture (cf.~Model 1 in Section~\ref{sec:model1}).}
 \label{fig:mse_mae_plot_endec_lstm}
   
\end{figure*}

The proposed models' performance with respect to MSE, MAE and per batch inference time is compared in Table~\ref{tab:results} and Fig.~\ref{fig:mse_mae_plot_atten_endec_lstm} - Fig.~\ref{fig:mse_mae_plot_endec_lstm}. Considering the overall target prediction precision of the models, one can find that the model--3, the global attention-based encoder-decoder LSTM surmounts all the models with an MSE value of \textbf{1.72} and MAE value of \textbf{5.34}. The alignment mechanism in the attention-based model is the key for these performance improvements when there is an unpredictable lagging between the predictor and target attributes, 
Hence, model--1 and model--3 have similar inference time ($\approx$ 13 $ms$), since the difference between the model is the only the attention layer added to model-3 that increases the number of parameters merely by 400. Although, such small increment in the trainable parameters has insignificant impact on the inference time due to parallel computation on GPU, but it enhances the prediction precision wrt MSE by $\approx31\%$ due to the added global attention mechanism. 

On the contrary, the model--2, bidirectional encoder-decoder LSTM triples the parameters compared to other two models, as it has to computes the hidden and cell state of the encoder subnetwork in both forward and backward directions (cf.Fig.~\ref{fig:bidirectional-lstm_cell}); thus, the inference time increases to $19~ms$ per batch. Even though we pay high cost for its bidirectional computation, unfortunately the prediction precision is not the best with MSE value of 3.94 and MAE value of6.39 compared to model--3. In other words, model--3 shows $\approx56\%$ and $\approx16\%$ performance improvement when compared to model--2 in terms of MSE and MAE, respectively.

Regarding a comparative study with the existing models, unfortunately, the literature has very inconstant way of reporting the results. Most of the literature do not provide a quantitative evaluation of individual target variables, rather they summarize the overall average performance. Thus, this work attempts to compare the best proposed model's performances with the best results found in a recent literature (cf. TCN~\cite{kirchgassner2020estimating}). It is found that this work's model--3 has improved the overall prediction by $>43\%$ and $>15\%$ when compared to TCN~\cite{kirchgassner2020estimating} with respect to MSE and MAE, respectively.

Considering all the above analysis and comparisons one can conclude that the proposed global attention-based encoder-decoder LSTM architecture shows robustness and higher performance for thermal stress prediction of permanent magnet synchronous motors.

\section{Conclusion}\label{conclusion}

The temperature prediction of PMSMs' internal and external components plays a vital role in myriad of industrial applications. However, it has been a very challenging task due to the complex internal structure of the PMSMs and heteroscedasticity nature of the components' thermal stress. To overcome the challenges, data-driven predictive models have been proposed that use externally measurable quantities to estimate the PMSMs' temperature. Nonetheless, still predicting all four target quantities, i.e., the temperature of \texttt{stator yoke, stator winding, stator tooth} and \texttt{permanent magnet}, simultaneously is cumbersome. Thus, some of the existing solutions attempt to predict single target value. In contrast, this work handles all four target predictions using LSTM-based EnDec architectures. It builds three unique models with varying complexities from a vanilla structure to global attention-based structure. The conducted exhaustive experiments show that the proposed global attention-based EnDec LSTM model exhibits better robustness and provides greater precision in predicting the PMSM's components' temperature.      

The future work is dedicated to further investigate the unstable performance of the bidirectional EnDec LSTM and to find more synthetic attributes that unveils useful interaction between the externally measurable attributes and enhance the prediction results.


%

\ifCLASSOPTIONcompsoc
  \section*{Acknowledgments}
\else
  \section*{Acknowledgment}
\fi

The authors would like to thank the the benchmark dataset organizers, the {Power Electrics and Drive Laboratory} at Paderborn University and Google for generously providing access to the high-performance computing (HPC) platform for machine learning via the Colab.

\ifCLASSOPTIONcaptionsoff
  \newpage
\fi



%



\bibliographystyle{IEEEtran}
\bibliography{reference.bib}

%







\end{document}